\documentclass[runningheads]{llncs}

\usepackage{eccv}

\usepackage[export]{adjustbox}
\usepackage{graphicx}
\usepackage{amsmath}
\usepackage{amssymb}
\usepackage{booktabs}
\usepackage{tabularx}
\usepackage{svg}
\usepackage[utf8]{inputenc}
\usepackage{CJKutf8}
\usepackage{rotating}
\usepackage{eccvabbrv}

\usepackage{graphicx}
\usepackage{booktabs}

\usepackage[accsupp]{axessibility}  % Improves PDF readability for those with disabilities.

\usepackage{hyperref}

\usepackage{orcidlink}
\usepackage{wrapfig}

\newcommand*\justify{%
  \fontdimen2\font=0.4em% interword space
  \fontdimen3\font=0.2em% interword stretch
  \fontdimen4\font=0.1em% interword shrink
  \fontdimen7\font=0.1em% extra space
  \hyphenchar\font=`\-% allowing hyphenation
}

\renewcommand{\texttt}[1]{%
  \begingroup
  \ttfamily
  \begingroup\lccode`~=`/\lowercase{\endgroup\def~}{/\discretionary{}{}{}}%
  \begingroup\lccode`~=`[\lowercase{\endgroup\def~}{[\discretionary{}{}{}}%
  \begingroup\lccode`~=`.\lowercase{\endgroup\def~}{.\discretionary{}{}{}}%
  \catcode`/=\active\catcode`[=\active\catcode`.=\active
  \justify\scantokens{#1\noexpand}%
  \endgroup
}

\begin{document}

% ---------------------------------------------------------------
\title{Spatial Colour Mixing Illusions as a Perception Stress Test for Vision-Language Models}

\titlerunning{Spatial Colour Mixing Illusions as a Perception Stress Test for VLMs}

% TODO FINAL: Replace with your author list. 
% Include the authors' OCRID for the camera-ready version, if at all possible.
\author{Nicoleta-Nina Basoc\inst{1}\orcidlink{0009-0001-2261-6944} \and
Adrian Cosma\inst{2}\orcidlink{0000-0003-0307-2520} \and
Emilian Radoi\inst{1}\orcidlink{0000-0002-1177-5288}}

% TODO FINAL: Replace with an abbreviated list of authors.
\authorrunning{N-N.~Basoc et al.}
% First names are abbreviated in the running head.
% If there are more than two authors, 'et al.' is used.

% TODO FINAL: Replace with your institution list.
\institute{National University of Science and Technology POLITEHNICA Bucharest, Bucharest, Romania \\
\and
IDSIA, The Dalle Molle Institute for Artificial Intelligence, Lugano, Switzerland
\email{nicoleta.basoc@stud.acs.upb.ro, adrian.cosma@idsia.ch, emilian.radoi@upb.ro}\\
}

\maketitle
\begin{figure}[hbt!]
    \centering
    \includegraphics[width=1\linewidth]{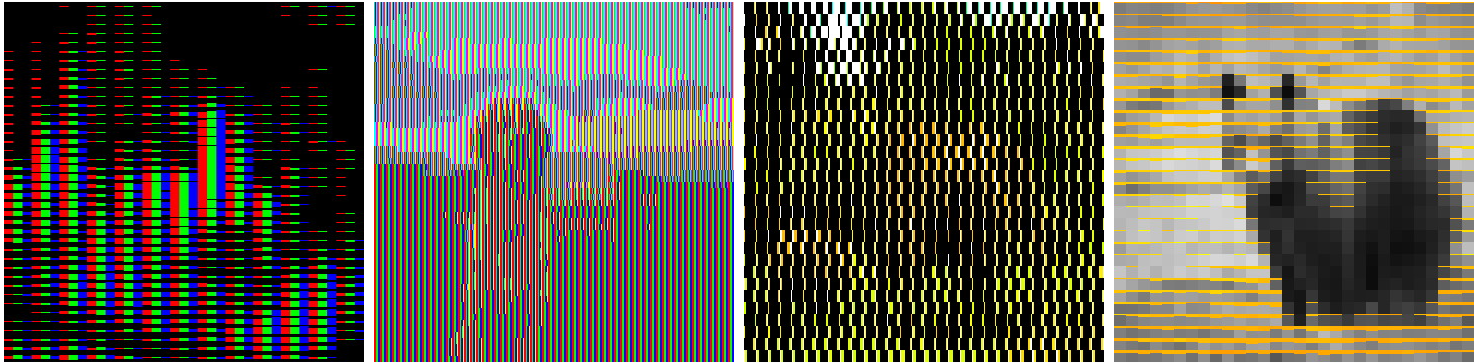}
    \caption[Caption for LOF]{What animals are shown in these examples? Readers are invited to distance themselves from the screen / zoom in and observe how their perception changes\protect\footnotemark[1]. Best viewed on a colour display.}
    \label{fig:teaser}
\end{figure}
\footnotetext[1]{Answer: \rotatebox[origin=c]{180}{From left to right: cat, dog, elephant, bear}}

\begin{abstract}
Vision-language models (VLMs) achieve strong benchmark results, yet can exhibit systematic perceptual weaknesses: structured, large changes to pixel values can cause confident yet nonsensical predictions, even when the underlying scene remains easily recognizable to humans. We study this gap using Spatial Colour Mixing, a programmatic family of colour distortions that overlays structured patterns (in both RGB and Ostwald colour systems) onto natural images. We introduce a framework of eight spatial colour mixing variants and evaluate nine VLMs across three model families on four datasets. Across models and datasets, accuracy degrades sharply with increasing distortion, and scaling the language model does not reliably mitigate the failure. In a human study with 61 participants on an animal recognition dataset, humans substantially outperform VLMs under the same distortions. Finally, we show that a simple human-inspired preprocessing step recovers a meaningful portion of performance for several distortion types, motivating perception-aware preprocessing and tool-use as practical strategies for improving VLM robustness.
\end{abstract}

\section{Introduction}
\label{sec:intro}
Generative Vision-Language Models (\emph{VLMs}) \cite{du2022survey,bordes2024introduction,li2025surveystateartlarge} have gained ubiquitous adoption for their visual understanding and reasoning capabilities across multiple domains~\cite{fu2025mmecomprehensiveevaluationbenchmark,liu2024mmbenchmultimodalmodelallaround}. Despite impressive benchmark performance, these models can fail in ways that are qualitatively surprising from a human perspective: small or structured changes to an image can cause confident but nonsensical outputs, even when the underlying scene remains easy for humans to interpret. We apply structured colour mixing distortions \cite{kitaoka2016spatialcolormixing} that substantially change pixel values while largely preserving the semantic content of the image. Humans can often recover the object identity quickly, and performance improves further when viewing the image "at a distance" (e.g., zooming out or stepping back), which reduces the influence of high-frequency details and makes the global shape more salient. Figure \ref{fig:teaser} reveals the phenomenon studied in this paper. In contrast, current VLMs can produce strikingly wrong answers on the same images.

A plausible reason for this mismatch is that human and machine perception are optimized for different objectives and operate under different constraints. Human vision is an active, constructive process that integrates partial evidence with strong knowledge priors \cite{article_attention,Gregory1997KnowledgeIP}. Furthermore, the interface theory of perception \cite{hoffman2015interface,prakash2021fitness} posits that humans and other animals perceive the world not as it is, but in such a way as to maximize evolutionary fitness. Similarly, VLMs perceive their inputs in such a way as to minimize their respective loss functions (i.e., contrastive or consistency losses \cite{tschannen2025siglip2multilingualvisionlanguage,simeoni2025dinov3}). It is then expected that there are stark differences in perception between the two entities. Illusions are the canonical tool for evaluating this mechanisms \cite{Gregory1997KnowledgeIP,article_understandin_hp_hm_illusions}. In the context of VLMs, illusions provide controlled perturbations that isolate low-level perception from high-level visual reasoning.

Recent work evaluates VLMs using illusion-based benchmarks~\cite{zhang2023grounding,shahgir2024illusionvqa,zhang2025illusionbenchlargescalecomprehensivebenchmark,mao2024evaluatingmodelperceptioncolor,rostamkhani2024illusoryvqabenchmarkingenhancing}. However, existing protocols often introduce subtle confounders: \textit{(i)} many illusion images are scraped from the internet and may be memorized; \textit{(ii)} common, well-known illusions reduce the evaluation to recognition of the illusion itself; \textit{(iii)} question wording and answer options can strongly constrain the hypothesis space, implicitly grounding perception in language \cite{wittgenstein21b}; and \textit{(iv)} visual prompting \cite{fu2024blinkmultimodallargelanguage} can change the task by introducing additional artifacts. 

We address this by focusing on an under-explored family of colour illusions\footnote{A colour illusion is a visual phenomenon in which the perceived colour of a region differs from its physical colour.}: \emph{Spatial Colour Mixing}~\cite{kitaoka2016spatialcolormixing,kitaoka2010brief}. Spatial colour mixing overlays structured stripe or grid patterns (under both RGB and Ostwald~\cite{granville1994color} colour systems) on top of real images, producing distortions that are large in pixel space yet often remain interpretable to humans. These transformations are programmatic: they can be applied to any image, enabling controlled evaluation over distortion strength without changing the underlying content. We treat spatial colour mixing as a perceptual confounder and measure how reliably VLMs preserve object identity as distortion increases.

In this work, we introduce a set of image perturbations for perceptual stress-testing of VLMs, built from eight spatial colour mixing transformations across two colour systems and multiple structural patterns, each parametrized by distortion degree. We evaluate nine VLMs across three widely used model families (Gemma~\cite{gemmateam2025gemma3technicalreport}, LLaVa~\cite{liu2023llava,liu2024llavanext}, Qwen3~\cite{bai2025qwen3vltechnicalreport}) across four different datasets (\emph{Animals}, \emph{Paintings}, \emph{Landmarks}, and the popular \emph{MME} \cite{fu2025mmecomprehensiveevaluationbenchmark} benchmark). 

We find that: \textit{(i)} scaling the language model does not mitigate the failure; \textit{(ii)} across a human study with 61 participants, there is a large gap between human and VLM performance on these distortions; \textit{(iii)} simple human-inspired preprocessing strategies (downscaling followed by bilinear upscaling and applying a blur box, simulating "stepping back" and "squinting"), recovers a meaningful fraction of performance on several illusion types. This suggests that VLMs could potentially overcome this limitation through tool-use \cite{wu2025vtoolr1vlmslearnthink}, but require the ability to recognize that their perception is unreliable. 

This work makes the following contributions:
\begin{enumerate}
    \item We propose a controlled family of \emph{Spatial Colour Mixing} distortions (eight variants across RGB and Ostwald systems) with a controllable distortion degree, enabling systematic stress testing without changing scene content. We make publicly available the distorted versions of each dataset under each illusion type. 
    
    \item We evaluate nine VLMs across three model families on four datasets under varying distortion strengths, quantifying how accuracy degrades as colour mixing intensity increases.
    
    \item We conduct a human study of 61 participants on the \emph{Animals} dataset to analyse the gap between human and VLM perception under the same distortions, and we release the dataset with associated human responses.
    
    \item We show that downscale-upscale and blurring preprocessing steps can substantially improve VLM accuracy on several illusion types, motivating tool-use and perception-aware preprocessing as practical countermeasures. We also evaluate \texttt{gpt-5-mini-2025-08-07} with \texttt{code-interpreter} tool use to test whether the model can recognize when its perception is unreliable, and find that the availability of tool use does not improve performance.
\end{enumerate}

\section{Related Work}
\label{sec:related}
\subsection{Vision-language models}

Recent years have seen rapid growth in VLM architectures \cite{du2022survey,bordes2024introduction,li2025surveystateartlarge}. The term \emph{VLM} is sometimes used broadly to include embedding models such as CLIP \cite{radford2021learningtransferablevisualmodels}. In this work, we use \emph{VLM} to mean generative language models that accept text and images as input and produce text. For broader background, we refer readers to existing surveys \cite{bordes2024introduction,xiao2026visionencoders}.

Most VLMs combine two components: a vision encoder and a language module \cite{xiao2026visionencoders}. These components are aligned using different training strategies, including contrastive objectives \cite{tschannen2025siglip2multilingualvisionlanguage,radford2021learningtransferablevisualmodels}, self-supervised vision pretraining \cite{jose2025dinov2,simeoni2025dinov3}, and alignment schemes that explicitly target an LLM interface \cite{chen2024internvl}. While language modeling has largely benefited from aggressively scaling model and data \cite{hoffmann2022training}, several studies suggest that scaling vision encoders can yield comparatively smaller gains. For instance, DINOv3 reports only modest improvements when increasing encoder size up to 7B parameters relative to much smaller variants \cite{simeoni2025dinov3}. In contrast, joint scaling of the vision encoder and the language model (e.g., PaLI \cite{chen2023pali-3} and PaLI-X \cite{chen2023pali-x}) has been reported to improve performance and to unlock additional capabilities at scale. Because large vision encoders require significant computational resources to train and deploy \cite{pmlr-v202-dehghani23a,simeoni2025dinov3}, many practical VLMs reuse a comparatively small, fixed encoder to reduce both training and inference overhead. Common choices include SigLIP \cite{zhai2023sigmoidlosslanguageimage} and CLIP ViT-L/14 \cite{radford2021learningtransferablevisualmodels,xiao2026visionencoders}.

Concrete examples follow this design pattern. The Gemma family \cite{gemmateam2025gemma3technicalreport} uses a shared 400M-parameter SigLIP vision encoder \cite{zhai2023sigmoidlosslanguageimage} across language-model scales, kept frozen during training. LLaVA-1.5 (7B, 13B) \cite{liu2024improvedbaselinesvisualinstruction} and LLaVA-1.6 (34B) \cite{liu2024llavanext} use a 300M-parameter CLIP ViT-L/14 encoder \cite{radford2021learningtransferablevisualmodels}, again typically frozen. The Qwen3-VL family \cite{bai2025qwen3vltechnicalreport} uses a fine-tuned SigLIP-2 encoder \cite{tschannen2025siglip2multilingualvisionlanguage}, with a 300M variant for smaller models and a 400M variant for larger ones.

\subsection{Evaluating VLM visual perception through illusions}

A large body of work evaluates the visual capabilities of VLMs using benchmarks such as MME~\cite{fu2025mmecomprehensiveevaluationbenchmark}, MathVista~\cite{lu2024mathvistaevaluatingmathematicalreasoning}, SEED-Bench~\cite{li2023seedbenchbenchmarkingmultimodalllms}, and MMBench~\cite{liu2024mmbenchmultimodalmodelallaround}. Such benchmarks cover diverse capabilities (e.g., recognition and visual reasoning), most are formulated as multiple-choice or heavily question-conditioned VQA.

A more controlled approach is to use visual illusions: illusions are appealing because they systematically perturb the mapping between the image and an observer's interpretation, probing the interaction between sensory cues and learned priors \cite{article_understandin_hp_hm_illusions}. However, illusion-based evaluations still vary substantially in what they measure and how strongly they rely on language scaffolding. In particular, existing benchmarks \cite{zhang2023grounding,shahgir2024illusionvqa,zhang2025illusionbenchlargescalecomprehensivebenchmark,mao2024evaluatingmodelperceptioncolor,rostamkhani2024illusoryvqabenchmarkingenhancing} differ along \textit{(i)} illusion type (e.g., optical vs. semantic), \textit{(ii)} data source (internet scrapes vs. programmatic transformations), and \textit{(iii)} evaluation protocol (multiple-choice or VQA vs. free-form answers). Each such axis introduces distinct confounders, for example, some VLMs might be bottlenecked by their ability to understand visual prompts~\cite{fu2024blinkmultimodallargelanguage} rather than their ability to perceive the image content.

The GVIL benchmark~\cite{zhang2023grounding}, explores VLM illusion perception by combining 16 colour and optical illusions in conjunction with questions. Their analysis emphasizes human-likeliness to the VLM answers, whereas our goal is more narrowly perceptual: given a real image in which a known object remains present, we evaluate whether a VLM can still identify that object under controlled colour perturbations. IllusionVQA \cite{shahgir2024illusionvqa} contains 374 cognitive illusions, collected from the internet, coupled with multiple-choice questions for VLM evaluation. Similarly, IllusionBench+ \cite{zhang2025illusionbenchlargescalecomprehensivebenchmark},  combines classic cognitive, real scenes Ishihara \cite{ishihara1918tests} and trap illusions into a 1051 image collection along with QA pairs. We instead, analyze underexplored colour distortions (i.e., RGB and Ostwald colour system-based spatial colour mixing) that can be programmatically applied to any image,  enabling controlled sweeps over distortion strength without changing scene content. RCID~\cite{mao2024evaluatingmodelperceptioncolor} combines contrast, stripe and filter colour illusions and evaluate VLM performance with different language prompting techniques, direct the VLM either on the pixel values or by assuming the role of a human to define of whether an illusion is present. The authors focused their evaluation on the model's colour understanding. Further, the IllusoryVQA\cite{rostamkhani2024illusoryvqabenchmarkingenhancing} and IllusionBench\cite{NEURIPS2024_a13ff984} datasets explore integrating well-known objects or concepts into sceneries using ControlNet~\cite{Zhang_2023_ICCV} and test the VLMs capabilities of shape recognition of pareidolia-type illusions for classification and OCR tasks. 

% In contrast, we focus on the perception despite colour alterations using a free-form evaluation protocol designed to minimize linguistic guidance, aiming to isolate how VLMs preserve object identity under controlled perceptual perturbations.

\subsection{Human visual perception}

Vision is central to human world understanding: it supports navigation, object interaction, and situation assessment. However, human perception is not well characterized as a passive copy of the external world. Instead, the percept is the outcome of an inference process that combines sensory inputs with prior experience, making systematic misperceptions (illusions) an expected failure mode of an efficient perceptual system \cite{article_understandin_hp_hm_illusions,Gregory1997KnowledgeIP}. A key aspect of human vision is that it is \emph{embodied} and tightly coupled to action and is thought to act like a functional (but not accurate) interface to the outside world \cite{hoffman2015interface,prakash2021fitness}. 

Moreover, human vision is also fundamentally temporal. High-acuity information is concentrated in the fovea, while the periphery provides lower-resolution but broader contextual signals. Perception therefore emerges from sequences of saccades and fixations that integrate information over time and across viewpoints \cite{articledemistifying}. This mechanism has no direct analogue in standard VLM pipelines, which typically encode a single static image in one forward pass. As a consequence, humans and VLMs can differ not only in \emph{what} information is available, but also in \emph{how} it is acquired and integrated.

Many accounts of illusion phenomena emphasize perceptual organization principles studied in Gestalt psychology: percepts reflect structured wholes rather than independent local parts. In particular, according to Van Geert et al.~\cite{articleprag}, by the gestalt law of Prägnanz, the principle of perceptual grouping, we perceive the simplest organised layout that corresponds to prior experience and knowledge. Strong evidence of such perceptual behaviour was registered by Fei-Fei et al.~\cite{10.1167/7.1.10}, which showed participants greyscale images flashed for short periods of time, ranging from 27ms to 500ms; participants were able to grasp the main object or scene type even for the shortest flashing durations. Rosenholtz~\cite{articledemistifying}, argues that the apparent mismatch between our richly experienced visual world and our poor access to its details is explained by efficient information-preserving encoding in peripheral vision, accounting for both vision’s successes and its failures.

Consequently, these studies suggests human vision is context, knowledge and experience-driven, naturally prone to illusions. As such, the human visual mechanism is different by construction from that of machine vision~\cite{Funke_2021}. Finally, prior work suggests that injecting simple human-motivated constraints into machine vision can sometimes improve alignment with human judgments on illusion tasks. For example, blur-based preprocessing has been reported to improve illusion recognition in certain settings \cite{rostamkhani2024illusoryvqabenchmarkingenhancing}, and peripheral-vision-inspired models such as texture tiling have been studied as approximations to human peripheral encoding \cite{articlettm,harrington2024cocoperiph}.

\section{Method}
\label{sec:method}
\subsection{Spatial colour mixing illusions}

In our work, we use spatial colour mixing illusions. Spatial colour mixing is a visual phenomenon grounded in trichromacy: the way the eye processes colour information through three distinct channels. This capability emerges from cone cells sensitive to three types of wavelengths, associated with red, green and blue. The signals from the cones are represented by the brain as various colours. This is the underlying principle of RGB screens, where simulating the trichromacy results in representing a vast colour spectrum. Spatial colour mixing illusions exaggerate this phenomenon by explicitly decomposing (i.e., in stripes or in a grid pattern) the colours of the image into the constituent colours of a colour system. To enhance the illusory effect, we adopt different variations of this technique \cite{kitaoka2016spatialcolormixing,kitaoka2010brief}. 

\begin{figure}[hbt!]
    \centering
     \includegraphics[width=1.0\linewidth]{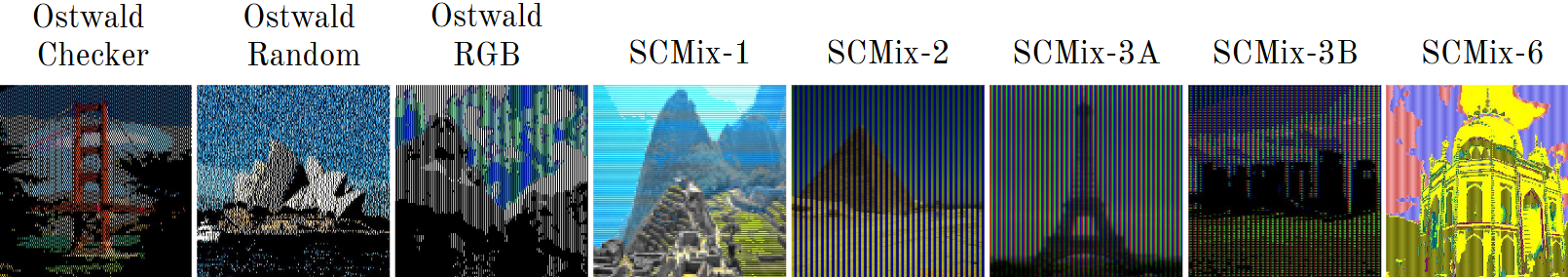}
    \caption{Examples of each of our 8 spatial colour mixing illusions.  Best viewed on a colour display.}
    \label{fig:example_illusion}
\end{figure}

We apply spatial colour mixing in five variants for the RGB system (\emph{SCMix-1}, \emph{SCMix-2}, \emph{SCMix-3A}, \emph{SCMix-3B}, \emph{SCMix-6}) and three variants in the Ostwald system \cite{granville1994color} (\emph{Ostwald RGB}, \emph{Ostwald Checker} and \emph{Ostwald Random}). Figure \ref{fig:example_illusion} showcases an example image under each type of colour illusion.

For the illusions in the RGB system, two types are constructed as coloured stripes superimposed on greyscale images vertically: \emph{SCMix-3A} is represented by three RGB coloured stripes, and \emph{SCMix-2}, which is represented by a combination of only two coloured stripes, $R+G$ and $B$. The other three variants encode chromatic information in stripes over greyscale image patches: \emph{SCMix-3B} is represented by RGB stripes of length proportional to their respective amount in the original image patch; \emph{SCMix-1} by only one stripe of colour averaged over the patch, and \emph{SCMix-6} represented by six coloured stripes (RGBCYM), also proportional to their amount in the original colours. 

The Ostwald colour system is based on the core Ostwald colour geometry \cite{granville1994color}: a colour can be decomposed into three components, black, white and hue, which form a perceived colour by adjusting their ratios. We use three forms of spatial mixing of the Ostwald colour system: \emph{Ostwald RGB}, \emph{Ostwald Checker} and \emph{Ostwald Random}. \emph{Ostwald RGB} represents vertical black, white and coloured RGB lines such that the sum of luminescence across three adjacent lines equals the original colour. \emph{Ostwald Checker} consists of representing the image as a grid where each block is divided into black, white and hue, in this particular order, such that their summed values equals the original perceived colour. \emph{Ostwald Random} follows the same principle as \emph{Ostwald Checker}, except the black and white are randomly alternated.

\begin{figure}[hbt!]
    \centering
     \includegraphics[width=0.85\linewidth]{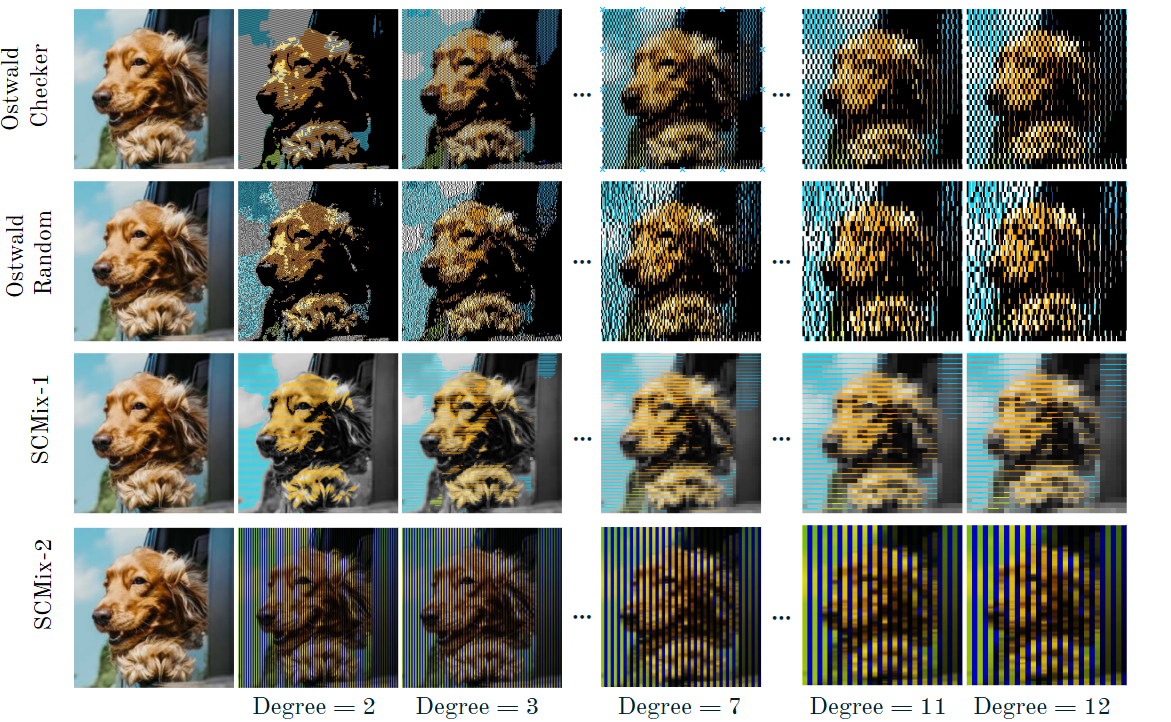}
    \caption{Examples of the effect of the distortion degree on four of our proposed colour mixing illusions. Best viewed on a coloured screen.}
    \label{fig:test}
\end{figure}

Each type of colour illusion can be parametrized by a degree of perturbation, which either corresponds to the width in pixels of the colour stripe (e.g., in \emph{SCMix-2}), or is proportional to the grid dimensions (e.g., in \emph{Ostwald Checker}). Examples are shown in Figure \ref{fig:test}.

\subsection{Datasets}
We employ three datasets that differ in the prompting specificity degree: from explicit prompts that anchor perception to predefined classes (\emph{Animals}), to less explicit (\emph{Artworks}) and free-form generation (\emph{Landmarks}). Each dataset contains inputs that are common on the internet and easy for the models to understand and correctly identify (i.e., common animals, famous paintings, popular landmarks). Furthermore, we also employ the popular MME~\cite{fu2025mmecomprehensiveevaluationbenchmark} benchmark for VLM performance evaluation across multiple axes. Before applying colour mixing onto the images across datasets, we resize all images to $360\times360$ resolution. We further describe the original datasets onto which we apply each illusion type.

\noindent \textbf{Animals.} 
This dataset consists of 1140 base images of 19 common animals, selected from the Kaggle Animals Image Dataset {\footnote{\url{https://www.kaggle.com/datasets/iamsouravbanerjee/animal-image-dataset-90-different-animals}, Accessed 4th of March 2026}}. The model is tasked to identify the animal present in the image. The prompt used for this dataset is: \emph{"What can you distinguish in this image? Please answer with one of the following: cat, dog, bee, eagle, cow, elephant, panda, tiger, horse, lion, penguin, bear, butterfly, flamingo, parrot, snake, bison, hippopotamus, hyena."}. This prompt provides class options to the model and linguistically grounds the visual perception. 

\noindent \textbf{Artworks.}
This dataset contains images of artworks curated from Kaggle Dataset Best Artworks of all Times {\footnote{\url{https://www.kaggle.com/datasets/ikarus777/best-artworks-of-all-time}, Accessed 4th of March 2026}}, containing 1951 images of paintings from 8 famous painters: Claude Monet, Edvard Munch, Leonardo da Vinci, Michelangelo, Pablo Picasso, Salvador Dali, Sandor Botticelli and Vincent van Gogh. The model is tasked to identify the artist of the given painting, and is prompted with the following question: \emph{"Who is the artist that painted the artwork shown in the image?"}. This prompt does not provide the available classes, but provides a hint that the image is an artwork. 

\noindent \textbf{Landmarks.}
This dataset is based on Google Landmarks Dataset v2 \cite{weyand2020googlelandmarksdatasetv2}, containing 3688 images of 15 notorious landmarks: Eiffel Tower, Tokyo Tower, Great Sphinx of Giza, Great Pyramid of Giza, Statue of Liberty, Louvre Pyramid, Golden Gate Bridge, Mount Rushmore, Burj Khalifa, Great wall of China, Stonehenge, Taj Mahal, Notre-Dame de Paris, Sydney Opera House and Machu Pichu. The models are tasked to identify the landmark. For this dataset, the models are prompted with an open question: \emph{"What can you distinguish in this image?"}, providing no hints about the image content.

\noindent \textbf{MME.}
Finally, we adopt the popular MME benchmark\cite{fu2025mmecomprehensiveevaluationbenchmark}, keeping the original structure and questions (1188 images along with 2376 questions). The benchmark contains 14 subtasks, testing the model capabilities in areas such as perception, text translation, celebrity recognition and others. For this dataset, we follow the official evaluation protocol: each image is associated with a description and two complementary questions, whose correct answer is either "yes" or "no". 

\subsection{Evaluation protocol}

We evaluate nine VLMs from the LLaVa \cite{liu2023llava,liu2024llavanext}, Gemma3 \cite{gemmateam2025gemma3technicalreport} and Qwen3 \cite{bai2025qwen3vltechnicalreport} families of models. From each family, we considered three model scales: \texttt{gemma-3-4b}, \texttt{gemma-3-12b} and \texttt{gemma-3-27b} for Gemma; \texttt{llava-1.5-7b}, \texttt{llava-1.5-13b} and \texttt{llava-v1.6-34b} for LLaVA; and \texttt{Qwen3-VL-4B}, \texttt{Qwen3-VL-8B} and \texttt{Qwen3-VL-30B} for Qwen3.

For an answer to be considered correct, the output generated by the model should contain the exact match of the correct class describing the animal / artwork / landmark that is contained in the image. Any other answer would be considered incorrect. We report the overall accuracy.

\subsection{Human evaluation}

We conduct a human evaluation on the \emph{Animals} dataset, on three degrees of intensity (2, 5 and 12). The subset was split into five complementary subgroups and were randomly assigned to 61 participants (43F/18M, university students, mean age 22). 
Twenty-nine participants had prescription glasses and five had contact lenses (and reported the prescriptions). None were colour blind.
%Participants were asked to report if they wear glasses, to report their prescription if applicable and if they are colour blind (none of the participants were colourblind, 29 of them have prescription glasses and 5 of them wore contact lenses).
Participants were instructed to look at the images from a distance of maximum 60cm from the screen, in adequate lightning, and to select from a multiple-choice field the animals they recognised. Images appear one after another on the screen, each for three seconds.

\subsection{Preprocessing via down/up rescaling and blurring}
\label{sec:preprocessing}

Previous works~\cite{rostamkhani2024illusoryvqabenchmarkingenhancing} have shown that VLMs can somewhat overcome pareidolia-type illusions by simulating human visual mechanisms: to better perceive the object in the image, human participants often reported distancing themselves from the screen, or see the image through their eyelashes (by squinting), to better perceive the image. These strategies plausibly reduce the influence of high-frequency stripe/grid patterns by averaging local neighbourhoods, making global shape cues more salient. VLMs can presumably use tools to obtain the same effect (e.g., write python code to manipulate the image), but requires that the model recognises that its perception is unreliable. Consequently, we test two preprocessing steps to attempt to alleviate the effect of spatial colour mixing illusions: downscaling and upscaling (D/U) and applying a box blur operation. We also evaluate \texttt{gpt-5-mini-2025-08-07} with \texttt{code-interpreter} to assess whether the model is capable of automatically determining if such additional preprocessing should be applied to distorted images.

\section{Experiments and Results}
\label{sec:exp-results}
\subsection{Main results}

\begin{figure}[hbt!]
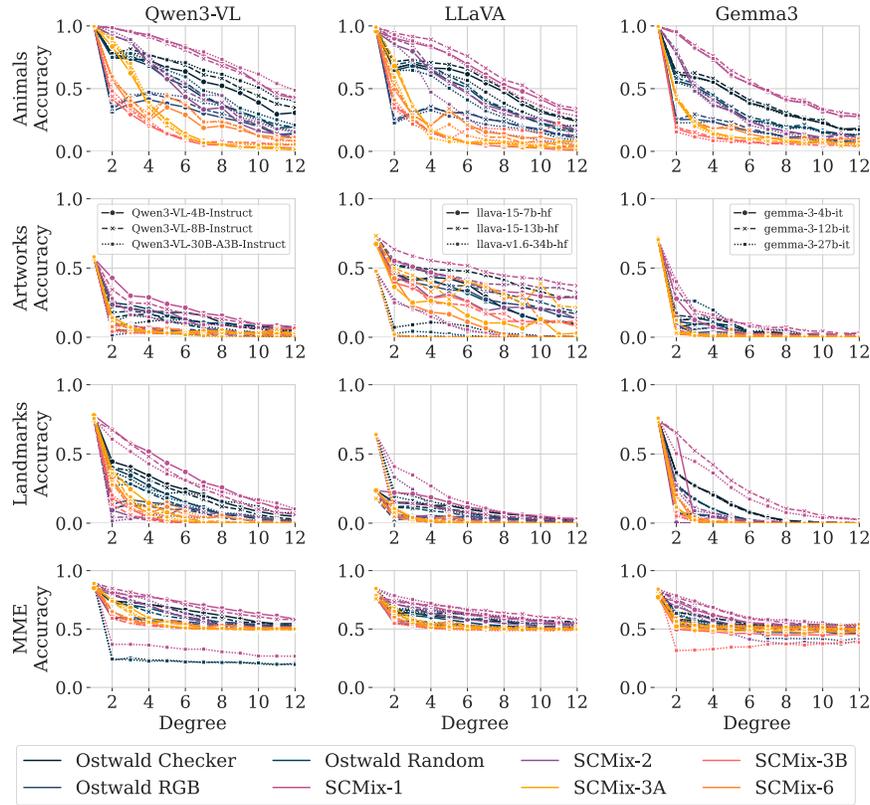

    \centering
    \includesvg[width=0.95\linewidth]{images/main_results_animals_vfinal.svg}
    \includesvg[width=0.95\linewidth]{images/main_results_artwork_vfinal.svg}
    \includesvg[width=0.95\linewidth]{images/main_results_landmarks_vfinal.svg}
    \includesvg[width=0.95\linewidth]{images/main_results_mme_vfinal.svg}
    \caption{Main results. Accuracy as a function of distortion degree for 9 VLMs across four datasets (\emph{Animals}, \emph{Artworks}, \emph{Landmarks}, \emph{MME}) under eight Spatial Colour Mixing variants. Columns group models by family (\emph{Qwen3-VL}, \emph{LLaVA}, \emph{Gemma3}); within each panel, line style indicates model scale and colour indicates the illusion type. Across datasets, accuracy degrades sharply even at low distortion degrees, and differences are driven more by model family than by language-model scale.}
    \label{fig:main-animals}
\end{figure}

In Figure \ref{fig:main-animals} we show task accuracy for each model and for each dataset across the four datasets we considered, across eight spatial colour mixing illusion types. Model performance drops dramatically after applying colour illusions even with the lowest distortion degree. In particular, for \emph{Animals}, arguably the easiest dataset, with models obtaining almost perfect performance in the undistorted case, performance drops to around 50\% after the first distortion degrees. This effect is even more pronounced for the other datasets. 

Across illusions types, we find that performance varies more by model family than by language model scale: this effect is most clearly seen for Gemma3, where the performance across scales is almost identical. For example, for \emph{Artworks}, LLaVa is the most robust model family, while Gemma3 and Qwen3-VL models are negatively impacted by the illusions, converging to 0\% accuracy. These effects might be attributed to each models' choice of vision encoder; we discuss this aspect in Section \ref{sec:vision-encoder}. 

For \emph{MME}, we find that for \texttt{Qwen3-VL-30B} and \texttt{gemma3-27b}, for certain illusion types, applying the distortions severely impacts impacts performance and results in non-sensical outputs for the model. For example, the Qwen3 model collapses and responds with \begin{CJK}{UTF8}{min}千里江山图\end{CJK}, a famous Chinese painting having stringent colours, or with a reference to Jackson Pollock. This indicates that the model might have overfitted and memorized such examples and does not attempt to perceive the given image for what it represents; this observation is in-line with the findings of Vo et al., \cite{vo2025vision}, which show that prior knowledge biases VLM predictions. \texttt{gemma3-27b}, on the other hand, for \emph{SCMix-3A}, does not provide a response and answers that "the image is a highly pixelated". In Table \ref{tables:examples_qualitatives} we show additional qualitative examples of model responses. These findings suggest that current VLMs struggle to efficiently process images under illusion distortions. 

\begin{table}[!htb]
    \centering
    \resizebox{0.9\linewidth}{!}{
        \begin{tabular}{p{0.15\linewidth} | p{0.30\linewidth} | p{0.15\linewidth} | p{0.30\linewidth} }
        \textbf{Fn/Image} & \textbf{Model Output} & \textbf{Fn/Image} & \textbf{Model Output} \\ \midrule

            %new_example
         SCMix-1
         \includegraphics[width=1.0 \linewidth, valign=t]{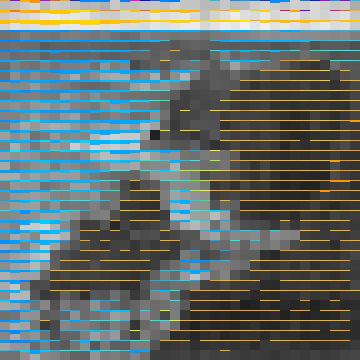} 
          & \texttt{LLaVA-7B}: "The artist who painted the artwork shown in the image is a dog."  
          &
          SCMix-2
        \includegraphics[width=1.0 \linewidth, valign=t]{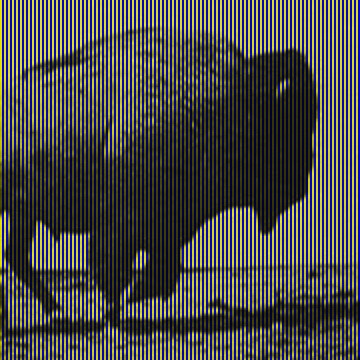}
        
          &\texttt{gemma3-27B}: "**hippopotamus** the image clearly depicts a hippopotamus."      \\ \midrule

            % %new_example
            SCMix-3A
            \includegraphics[width=1.0 \linewidth, valign=t]{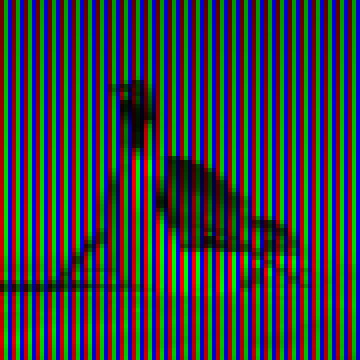}
              & \texttt{gemma3-12B}: "cat. it's cleverly hidden within the striped background!" 
              &
              Ostwald Random
            \includegraphics[width=1.0 \linewidth, valign=t]{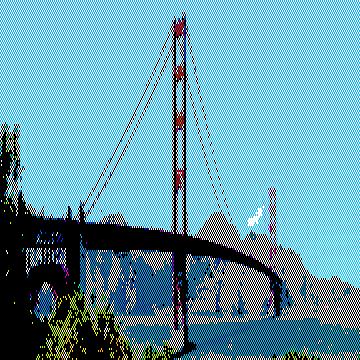}
            
              & \texttt{Qwen3-VL-30B}: "\begin{CJK}{UTF8}{min} 千里川土手\end{CJK}, a location in japan."    \\ \midrule

            % %new_example
            Ostwald Random
            \includegraphics[width=1.0 \linewidth, valign=t]{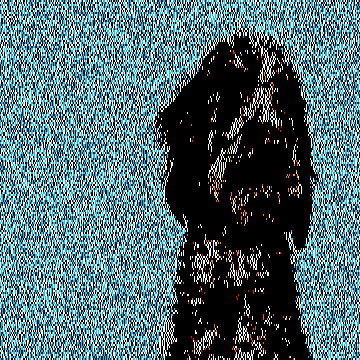}
            
              & \texttt{LLaVA-NEXT-34B}: "the image appears to be a low-resolution or pixelated image of a person, not an animal. it's difficult to discern specific details due to the quality of the image."  &
              SCMix-3B
            \includegraphics[width=1.0 \linewidth, valign=t]{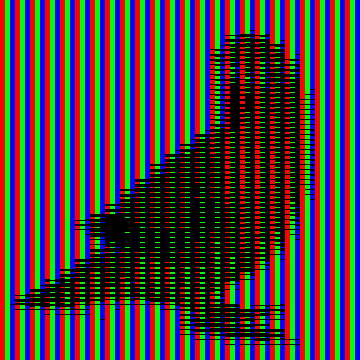}
            
              & \texttt{gemma3-27B}:  "**bear** the image is an optical illusion where the shape of a bear is hidden within a colorful, striped background. it takes a moment to focus and perceive the animal's form."    \\ \midrule
              
             %new_example
             Ostwald Checker
            \includegraphics[width=1.0 \linewidth, valign=t]{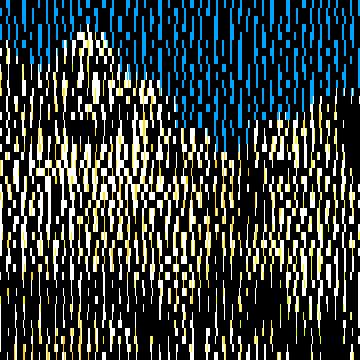}
              &\texttt{Qwen3-VL-30B}: "this image is a pixelated, low-resolution depiction of the statue of liberty."
              % the iconic statue is recognizable by its distinct shape, including the crown with its spikes and the torch held aloft. the image uses a limited color palette, primarily black, white, and blue, which creates a blocky, digital appearance. the blue areas likely represent the sky or water in the background, while the white and black pixels form the structure of the statue itself. the overall effect is a stylized, abstract representation of this famous landmark.    
              &
              Ostwald Checker
            \includegraphics[width=1.0 \linewidth, valign=t]{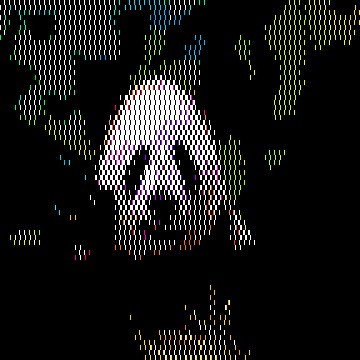}
            
               &\texttt{gemma3-12B}: "based on the image, i can distinguish a **skull**."  \\ 
              % \midrule
        \end{tabular}}
        \caption{Representative distorted inputs and corresponding model outputs illustrating non-robust behaviour beyond performance degradation: e.g., off-task responses (animal names as artists), confident misidentifications, "illusion-aware" descriptions that still miss the correct concept, and occasional off-distribution generations (including foreign-language text).  Best viewed on a colour display.}
        \label{tables:examples_qualitatives}
\end{table}

\begin{figure}[hbt!]
    \centering
    \includesvg[width=0.85\linewidth]{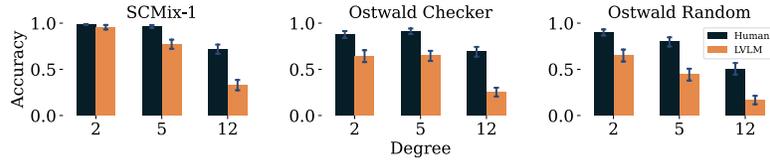}
    \caption{Mean per-image accuracy on the Animals dataset for three distortion degrees (2, 5, 12) and three representative illusion types (\emph{SCMix-1}, \emph{Ostwald Checker}, \emph{Ostwald Random}). Bars compare human responses to the aggregated performance of the nine evaluated VLMs.}
    \label{fig:human_eval}
\end{figure}

\subsection{Human evaluation}

In Figure \ref{fig:human_eval} we show average accuracy of human participants compared to the aggregated accuracy of the VLMs in our experimental setup.  There is a substantial performance gap between VLMs and human participants. We use Fleiss Kappa\cite{articlefleiss} to measure inter-rater agreement and obtain a $\kappa$-score of 0.748. We find that humans are more robust to spatial colour mixing illusions, obtaining higher performance than VLMs, and the task performance degrades much slower than VLMs with increasing the distortion degree. This is evidence that VLMs process images differently under spatial colour mixing perturbations. Other works~\cite{bavaresco2024modelling} found that VLMs output brain-aligned representations when tested on natural images and on high level semantic concepts. However, it is clear that in terms of low-level perception, models qualitatively differ from human processing: for images that are easy for human to process, VLMs struggle \cite{fu2024blinkmultimodallargelanguage}.

\begin{figure}[hbt!]
    \centering
    \includesvg[width=0.85\linewidth]{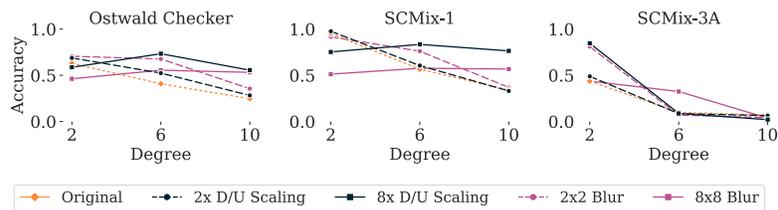}
    \caption{Results for \texttt{gemma-3-12b} on the \emph{Animals} dataset for preprocessing images using human-inspired low-pass filtering procedures: downscaling followed by upscaling (D/U) and applying box blurring to the distorted images using \emph{Ostwald Checker}, \emph{SCMix-1}, \emph{SCMix-3A}. Aggressive low-pass filtering improves performance on \emph{Ostwald Checker} and \emph{SCMix-1}, but does not affect illusions with multiple lanes (e.g., \emph{SCMix-3A}).}
    \label{fig:resize-animals}
\end{figure}

\subsection{Human-inspired low-pass preprocessing}

Following Section~\ref{sec:preprocessing}, we apply two lightweight operations to the \emph{distorted} images before feeding them to the model: \textit{(i)} downscaling followed by bilinear upscaling (D/U), and \textit{(ii)} box blurring. We treat these as external preprocessing steps rather than architectural changes, and we do not assume the model can reliably decide when to apply them.

Figure~\ref{fig:resize-animals} shows results on \emph{Animals} for three representative illusion types (\emph{Ostwald Checker}, \emph{SCMix-1}, \emph{SCMix-3A}). Since the Gemma family of models obtained the lowest performance on this dataset (see Figure \ref{fig:main-animals}), we use \texttt{gemma-3-12b} for this experiment. We observe that more aggressive low-pass filtering can substantially improve accuracy for \emph{Ostwald Checker} and \emph{SCMix-1} across distortion degrees, partially recovering performance under moderate-to-strong distortions: 8$\times$ D/U factor seems to have the largest impact as it increases model performance by more than 30\% for distortion degree 10. In contrast, \emph{SCMix-3A} shows little to no improvement, as multi-lane stripe patterns are not removed by simple averaging and may continue to dominate the encoder representation.

\begin{wrapfigure}{r}{0.40\columnwidth}
    \vspace{-0.5\baselineskip}
  \centering
  \includesvg[width=\linewidth]{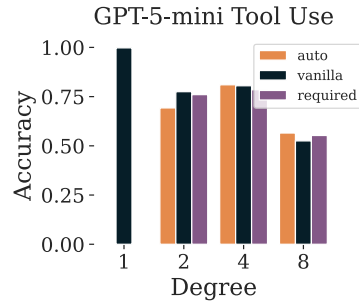}
  \caption{Results for tool use on a subsample of \emph{Animals} using \emph{Ostwald Checker} perturbation using \texttt{gpt-5-mini-2025-08-07}  and \texttt{code-interpreter}. Here, "auto" refers to optional tool use. Degree "1" refers to original images.}
  \label{fig:inline-gpt}
  \vspace{-2\baselineskip}
\end{wrapfigure}

To evaluate whether models can decide to use the low-pass filtering preprocessing on their own, we use \texttt{gpt-5-mini-2025-08-07} with \texttt{code-interpreter} tool use~\cite{qu2025tool} in three variations: the standard version (vanilla), a version with optional \texttt{code-interpreter} (auto), and a version where tool use is mandatory (required). We find that the availability of tools does not improve performance (Figure \ref{fig:inline-gpt}), implying that the model fails to recognize when its perception is unreliable.

\subsection{Effect of the vision encoder}
\label{sec:vision-encoder}

\begin{figure}[hbt!]
    \centering
    \includesvg[width=0.85\linewidth]{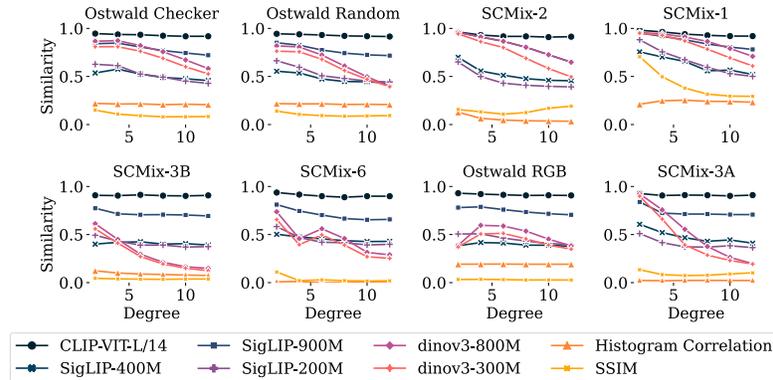}
    \caption{Average cosine similarities between original images and their corresponding illusions based on mean-pooled embeddings generated by multiple vision encoders (CLIP-VIT, SigLIP and DINOv3). We also show SSIM and histogram correlation across distortion degrees.}
    \label{fig:embedding_similarity}
\end{figure}

In Figure~\ref{fig:embedding_similarity} we show how encoder embeddings change as spatial colour mixing distortions intensify, by computing cosine similarity between an image and its distorted version using mean-pooled features from several vision backbones. Two patterns stand out: \textit{(i)} scale has limited impact within an encoder family: for example, SigLIP variants behave similarly to each other and DINOv3 \cite{simeoni2025dinov3} variants behave similarly to each other. And \textit{(ii)} the training objective dominates across families: CLIP/SigLIP similarities often remain unusually high and relatively flat with increasing distortion, whereas DINOv3 similarities drop more consistently with distortion degree. This aligns with broader findings of VLM design choices, which find that encoder pretraining and alignment strategy contribute more than scaling an strong backbone \cite{xiao2026visionencoders,kar2024brave}.

The DINOv3 trend suggests that self-supervised features retain mid-level structure disrupted by these distortions, making them more diagnostic of severity in representation space. This does not imply changing CLIP for DINO in generative VLMs, since most VLMs rely on language-aligned encoders and consume token-level features through a connector rather than mean-pooled embeddings. Still, DINO features are becoming viable in VLM pipelines: Jose et al., \cite{jose2025dinov2} explicitly aligns DINOv2 to language for open-vocabulary use, and BRAVE \cite{kar2024brave} consolidates multiple frozen encoders to broaden VLM visual evidence. Practically, Figure \ref{fig:embedding_similarity} supports findings consistent with perception-focused evaluations \cite{fu2024blinkmultimodallargelanguage}: scaling alone is not enough for robust perception, making hybrid designs that combine CLIP/SigLIP semantics with DINOv3 perceptual biases a promising direction for improving robustness to structured distortions like spatial colour mixing.

\section{Conclusions}
\label{sec:conclusions}
We introduced Spatial Colour Mixing as a controlled family of programmatic colour distortions for stress-testing VLM perception, spanning 8 variants across RGB and Ostwald systems and a tunable distortion degree. Across nine VLMs and four datasets, accuracy collapses rapidly even at low distortion strengths, and increasing language-model scale does not reliably mitigate this brittleness. A human study on \emph{Animals} shows a substantial robustness gap: humans retain object identity under the same perturbations far more consistently than current VLMs, indicating a mismatch in how low-level visual evidence is encoded. 

We also found that simple, human-inspired low-pass preprocessing (downscale-upscale or blurring) can recover a meaningful fraction of performance for several illusion types, but that providing tool use does not automatically yield the same benefit: models often fail to recognize when their perception is unreliable. Finally, embedding analyses suggest that encoder inductive bias matters more than encoder scale, with DINOv3 features showing more distortion-sensitive trends than CLIP/SigLIP encoders. These results motivate perception-aware preprocessing and hybrid visual encoders, as well as mechanisms for uncertainty-aware tool invocation, as practical directions to improve VLM robustness to colour illusions.

% TODO FINAL VERSION
\section*{Acknowledgements}
\label{sec:ack}
This research was supported by the project "Romanian Hub for Artificial Intelligence - HRIA", Smart Growth, Digitization and Financial Instruments Program, 2021-2027, MySMIS no. 351416.

\bibliographystyle{splncs04}
\bibliography{refs}
\end{document}